\documentclass{article}

\usepackage{graphicx}
\graphicspath{{}{graphs/}}

\usepackage[preprint]{neurips_2025}

\usepackage[utf8]{inputenc} 
\usepackage[T1]{fontenc}    
\usepackage{hyperref}       
\usepackage{url}            
\usepackage{booktabs}       
\usepackage{amsfonts}       
\usepackage{nicefrac}       
\usepackage{microtype}      
\usepackage{xcolor}         
\usepackage[utf8]{inputenc}
\usepackage{xcolor}
\usepackage[skins,breakable]{tcolorbox}

\definecolor{promptbg}{HTML}{F9F9F9}
\definecolor{reflectionbg}{HTML}{E8F4F8}
\definecolor{errorbg}{HTML}{FDECEC}

\tcbset{
    trajectorybox/.style={
        colback=promptbg,
        colframe=gray!50,
        boxrule=0.5pt,
        arc=4pt,
        left=6pt,
        right=6pt,
        top=6pt,
        bottom=6pt,
        fonttitle=\bfseries,
        coltitle=black,
        colbacktitle=gray!15,
        before upper={\raggedright},
        breakable
    },
    reflectionbox/.style={
        colback=reflectionbg,
        colframe=blue!30,
        boxrule=0.5pt,
        arc=2pt,
        left=4pt,
        right=4pt,
        top=4pt,
        bottom=4pt,
        before upper={\raggedright},
        breakable
    }
}

\title{From Hallucination to Structure Snowballing: The Alignment Tax of Constrained Decoding in LLM Reflection}

\author{%
  Hongxu Zhou \\
  Erasmus Mundus Master's in Language and Communication Technologies\\
  Saarland University\\
  Saarbrücken, Germany \\
  \texttt{hozh00003@stud.uni-saarland.de} \\
}

\begin{document}

\maketitle

\begin{abstract}
  Intrinsic self-correction in Large Language Models (LLMs) frequently fails in open-ended reasoning tasks due to ``hallucination snowballing,'' a phenomenon in which models recursively justify early errors during free-text reflection. While structured feedback can mitigate this issue, existing approaches often rely on externally trained critics or symbolic tools, reducing agent autonomy. This study investigates whether enforcing structured reflection purely through \texttt{Outlines}-based constrained decoding can disrupt error propagation without additional training. Evaluating an 8-billion-parameter model (Qwen3-8B), we show that simply imposing structural constraints does not improve self-correction performance. Instead, it triggers a new failure mode termed ``structure snowballing.'' We find that the cognitive load required to satisfy strict formatting rules pushes the model into formatting traps. This observation helps explain why the agent achieves near-perfect superficial syntactic alignment yet fails to detect or resolve deeper semantic errors. These findings expose an ``alignment tax'' inherent to constrained decoding, highlighting a tension between structural granularity and internal model capacity in autonomous workflows. Code and raw logs are available in the GitHub repository: \url{https://github.com/hongxuzhou/agentic_llm_structured_self_critique}.
\end{abstract}

\section{Introduction}

Recent studies have explored frameworks that instruct large language models to correct their own outputs. The Reflexion framework established a foundation for ``verbal reinforcement learning'' proposed by \citep{shinnReflexionLanguageAgents2023a}. In this setup, an agent generates natural language reflections after a failed task attempt, stores them in an episodic memory, and adjusts its behaviour in subsequent trials. This approach has shown effectiveness in environments with clear external grounding, such as code generation tasks.

However, researchers have identified significant limitations in intrinsic self-correction. Language models often fail to reliably correct reasoning errors without external validation signals, and their overall performance can even decrease after attempting to self-correct \citep{huangLargeLanguageModels2024}. \citet{zhangHowLanguageModel2023} argued that a key mechanism behind this failure was hallucination snowballing. During free-text reflection, models tend to generate self-consistent but false explanations for early errors. This process consolidates the original mistake within the reasoning chain rather than resolving it.

Structured feedback presents a potential solution to this issue. Research indicates that typed, formatted feedback performs better than free-text reflection. For example, the REFINER framework significantly improves reasoning task performance by using an external critic to provide structured feedback formats \citet{paulREFINERReasoningFeedback2023}. Yet, these structured approaches rely heavily on external tools, separately trained critic modules, and/or symbolic solvers. This reliance undermines the autonomy of the language model's intrinsic self-correction. A critical gap remains regarding whether structured feedback formats can be enforced purely through grammar-constrained decoding during generation, without depending on external training or tools.

To address this gap, this study investigates the following research questions:

\begin{enumerate}
    \item Can enforcing structured reflection through grammar-constrained decoding independently disrupt hallucination snowballing and improve self-correction success rates in open reasoning tasks?
    \item How does the error propagation mechanism change when a model conducts intrinsic reflection under strict structural constraints?
    \item Is there any tension exists between a model's internal capacity and the granularity of structural constraints during self-correction?
    
\end{enumerate}

Contrary to initial expectations, this study finds that simply introducing structural constraints does not improve performance. Instead, it introduces a new vulnerability, which we term ``structure snowballing''. While constrained decoding successfully suppresses divergent semantic hallucinations, it forces the model into formatting traps and death loops. Smaller models, such as the Qwen3-8B \citep{yangQwen3TechnicalReport2025} model used in this study, expend substantial cognitive effort to meet strict classification and syntactic rules. As a result, the model achieves perfect superficial syntactic alignment but entirely misses deep semantic errors.

This paper makes three main contributions. First, it provides empirical evidence that intrinsic structured self-correction via constrained decoding introduces an alignment tax when deployed without external training. Second, it defines and demonstrates structure snowballing as a theoretical counterpart to hallucination snowballing, revealing a new failure mode in constrained self-reasoning. Finally, it explores the practical boundaries of deploying complex agent logic on smaller open-weight models.
\section{Related Work}
\subsection{Self-Correction } 
Recent studies explore frameworks that instruct language models to correct their own outputs. The Reflexion framework introduces verbal reinforcement learning, where an agent generates natural language reflections after a failed task attempt, stores them in episodic memory, and adjusts its behaviour in subsequent trials \cite{shinnReflexionLanguageAgents2023a}. Similarly, methods like Self-Refine use iterative feedback cycles to improve initial outputs across different tasks \cite{madaanSelfRefineIterativeRefinement2023}.
However, researchers have identified significant limitations in intrinsic self-correction. Language models often fail to correct reasoning errors reliably without external validation signals, and their performance can even decrease after attempting to self-correct \cite{huangLargeLanguageModels2024}. The primary bottleneck lies in error detection rather than error correction. Models show limited capability in finding their own mistakes within a reasoning chain \cite{tyenLLMsCannotFind2024}. When the self-correction process is decomposed into error detection, localisation, and correction, evidence indicates that unguided models struggle to accurately identify where the reasoning process breaks down \cite{liDecomposingLLMSelfCorrection2025}. This highlights the necessity for explicit guidance during the reflection phase.
\subsection{Structured Feedback in Reasoning}
Providing models with structured, specific feedback addresses the shortcomings of generic natural language reflections. The REFINER framework demonstrates that semi-structured feedback, which explicitly identifies error types and their positions, improves intermediate reasoning steps significantly more than free-form text or scalar rewards. Effective structured feedback requires a clear categorisation of reasoning failures. Frameworks like ROSCOE \citep{golovnevaROSCOESuiteMetrics2023} provide a taxonomy for evaluating step-by-step reasoning, defining specific error types such as logical inconsistencies, missing steps, and hallucinations. Furthermore, research into multi-aspect feedback suggests that separating feedback generation into specific error categories prevents the model from generating vague responses and improves the overall quality of the corrections.
Despite these benefits, current methods for generating structured feedback are computationally expensive. Implementing frameworks like REFINER requires training an independent critic model using annotated datasets \citep{paulREFINERReasoningFeedback2023}. Other approaches depend on complex external tools or API calls to verify logic. This creates a research gap regarding how to generate reliable, structured feedback during inference without the cost of supervised fine-tuning or external model dependency.
\subsection{Constrained Decoding}
Constrained decoding provides an engineering solution to enforce structured outputs without additional training. \texttt{OUTLINES}, developed by \citet{willardEfficientGuidedGeneration2023} is one of the most commonly used tool for grammar constraind decoding generation. It is based on finite-state machines to guide text generation by constructing an index over the vocabulary of the language model, which guarantees that the output conforms strictly to predefined regular expressions or context-free grammars. This approach is model-agnostic and adds minimal computational overhead to the token generation process.

Applying hard constraints through tools like Outlines allows developers to enforce predefined error taxonomies during the reflection phase. By shifting the formatting requirements from the training layer to the decoding layer, this mechanism reduces the cognitive load on smaller models. It forces the model to diagnose errors into specific categories, bridging the gap between the need for structured reasoning feedback and the high cost of training dedicated critic models.

\section{Methodology}

\subsection{A 4+1 Taxonomy of Multi-hop Reasoning Errors}

To systematically evaluate and correct failures in multi-hop reasoning, we propose a 4+1 taxonomy of errors. The first four categories represent failures in cognition and reasoning, while the final category addresses formatting alignment. 

\begin{enumerate}
    \item \texttt{RETRIEVAL\_FOCUS:} the model identifies the wrong paragraph or sentence initially.
    \item \texttt{BRIDGE\_FAILURE:} the model locates the correct initial information but fails to extract the necessary bridge entity to transition to the next reasoning step.
    \item \texttt{HALLUCINATION:} the model generates facts or details that are not present in the provided context \citep{golovnevaROSCOESuiteMetrics2023}
    \item \texttt{INFERENCE\_ERROR:} the model successfully locates information but makes a logical deduction or arithmetic calculation error \citep{golovnevaROSCOESuiteMetrics2023}
    \item \texttt{FORMATTING\_MISMATCH:} the model derives the correct conclusion but fails to output it using the required formatting constraints
\end{enumerate}

These categories map to an information flow funnel representing the sequential cognitive steps of an agent. The steps include addressing information $\rightarrow$ linking entities $\rightarrow$ respecting knowledge boundaries $\rightarrow$  performing logical deduction $\rightarrow$ formatting the response. This flow ensures the error categories are mutually exclusive and collectively exhaustive.

We limit the taxonomy to five categories as a specific design choice to balance diagnostic granularity with model capacity. \citet{golovnevaROSCOESuiteMetrics2023} have defined up to ten fine-grained reasoning errors. However, \citet{nathaniMAFMultiAspectFeedback2023} note that multiple error categories simultaneously presents a significant challenge for language models with limited context lengths: Supplying a multitude of error types in a single instance burdens the model and degrades the quality of the generated feedback. We therefore restrict the taxonomy to five categories to prevent model's cognitive overload. This ensures that the 8B-parameter model focuses its attention on the most critical bottlenecks in multi-hop reasoning.

\subsection{First Attribution Strategy}

Reasoning failures in large language models rarely occur in isolation. An initial mistake during the early stages of generation often triggers a sequence of subsequent errors. Regarding it, we implement an upstream-first attribution strategy to address this issue. When a reasoning trajectory fails, the reflector is required to identify and classify only the first fatal error that occurred in the sequence.

This, according to \citep{zhangHowLanguageModel2023}, interrupts the hallucination snowballing effect. Language models naturally tend to commit to their early choices. If an early mistake is made, the model will often over-commit to it and produce subsequent hallucinations or flawed logic to justify the initial error. Correcting a downstream logical error without addressing the upstream retrieval failure is therefore ineffective.

Our strategy aligns directly with the annotation principles established in the BIG-Bench Mistake dataset by \citet{tyenLLMsCannotFind2024}. In that dataset, each generated trace is explicitly annotated with the first logical error, and any subsequent errors are ignored because they are typically dependent on the original mistake. We enforce this priority during the reflection stage so that our structured feedback precisely targets the root cause of the reasoning failure rather than its downstream symptoms.

\subsection{Logic-Guided Reflection via Constrained Decoding}
We construct our logic-guided self-correction framework upon a ternary architecture consisting of an Actor, an Evaluator, and a Reflector. Following the standard self-reflection paradigm \citep{shinnReflexionLanguageAgents2023a}, the Actor ($M_a$) acts as a policy function that generates a reasoning trajectory $\mathrm{Trajectory} \sim \pi_\Theta(y|x)$ given the input context. The Evaluator ($M_e$) assesses the final generated answer against the ground truth to provide a binary reward signal. When a trajectory fails, the Reflector ($M_r$) diagnoses the root cause of the error based on the failed trajectory and an Episodic Memory $M$, which stores previously accumulated correction rules.

Traditional self-correction frameworks often rely on free-text reflections or require training a specialised, computationally expensive critic model to generate structured feedback. To achieve high-quality structured feedback without the substantial cost of fine-tuning, we implement logic-guided reflection via constrained decoding. Instead of relying on prompt engineering alone to enforce formatting, we use the \texttt{Outlines} library to translate a predefined Pydantic schema-comprising our 5-class \texttt{Error Type } and a string \texttt{Correction\_Rule} into a finite-state machine (FSM).

From a machine learning mechanism perspective, this FSM operates directly at the logits level during the autoregressive decoding process. At each time step, the mechanism computes an un-normalised conditional distribution by applying a dynamic boolean mask that restricts the support of the original next-token distribution. This masking strictly forces the probabilities of any tokens that violate the JSON syntax or fall outside our predefined taxonomy enumeration to zero. As a result, we mathematically guarantee 100\% schema adherence and formatting compliance. This allows an off-the-shelf 8B-parameter model to reliably perform targeted, structured error attribution comparable to a fine-tuned critic model.

Once the Reflector successfully generates a valid structured output containing the exact error type and correction rule, this JSON string is appended to the Episodic Memory $M$. This updated episodic memory serves as a direct diagnostic context for the Actor's subsequent trial, effectively closing the loop and preventing the repetition of the identified logical misstep.

The full definitions of the taxonomy, system prompts including few-shot demonstrations, and decoding implementation details are provided in Appendix.

\section{Experimental Design}

\subsection{Dataset and Pre-processing}
This study investigates reasoning models using the HotpotQA dataset in its distractor setting \citep{yangHotpotQADatasetDiverse2018}. This setting provides the model with two gold paragraphs alongside eight distracting paragraphs to test information extraction and focus. An initial pilot test on 1,000 samples revealed that the Qwen3-8B model correctly answered 631 instances on its first attempt. \citet{minCompositionalQuestionsNot2019} demonstrated that modern language models often bypass multi-hop reasoning requirements by exploiting dataset shortcuts. This high initial success rate indicates that classic datasets may experience saturation, as many questions no longer force modern models to execute self-correction. We must therefore clean the data to isolate genuinely challenging samples.

Traditional evaluation metrics for question answering, such as Exact Match (EM) and F1 scores, suffer from severe brittleness. For example, if the gold answer is the digit ``4'' and the model outputs the word ``four'', strict string matching incorrectly marks the answer as a failure. This rigidity produces a high rate of false negatives. To address this issue, we employ an LLM-as-a-judge pipeline to assess semantic equivalence and entailment \citep{zhengJudgingLLMasaJudgeMTBench2023}. We use the Qwen3-8B model as a judge to verify whether the output semantically matches the gold answer. This process filters out simple questions and eliminates false negatives caused by superficial formatting differences.

\subsection{Two-Pool Evaluation Strategy}
We divide the filtered dataset into two distinct evaluation pools to assess the efficiency and the upper capability limits of structured reflection. \citet{shinnReflexionLanguageAgents2023a} noted that standard Reflexion often struggles with open-ended reasoning tasks. By separating the data, we prevent evaluation bias caused by reasoning shortcuts and specifically target difficult instances. 

Pool A consists of 55 samples that the baseline Reflexion framework successfully solved between the second and fifth trials. For this group, the primary evaluation metric is Average Trajectories (AT). We use AT to measure whether structured constraints can guide the model to the correct answer in fewer steps. Pool B comprises 45 samples where the baseline model completely failed. These ones often fell into continuous hallucination loops and failed to finish within the maximum allowed steps \citep{zhangHowLanguageModel2023}. For Pool B, the primary evaluation metric is the Success Rate (SR), which determines if structured reflection can break the model out of failure states.

\subsection{Baselines and Implementation Details}
We compare our proposed method against a standard Reflexion framework to isolate the effect of structural decoding. The Control group utilises the baseline Reflexion architecture, where the internal reflector generates self-correction feedback in free natural language \citep{shinnReflexionLanguageAgents2023a} The Treatment group employs the Logic-Guided Reflexion architecture outlined in Section 3.3. Both groups share identical retrieval settings and verification mechanisms.

To minimise random variance during the reasoning process, the generation temperature is set to 0.1 and the maximum generation length is restricted to 1024 tokens across all trials. Due to resource constraints, we deploy the Qwen3-8B model for both generation and evaluation tasks. Maintaining a uniform model for both the Actor and Evaluator roles prevents capability mismatches and mitigates intra-model scoring bias. Furthermore, we strictly limit the evaluator to verifying semantic equivalence, avoiding subjective quality ranking. Finally, the upstream-first attribution strategy defined in Section 3.2 is enforced uniformly during the treatment condition to disrupt hallucination snowballing.

\section{Results and Interpretation}
\subsection{Overall Efficacy and the Limits of Intrinsic Constraints}

\begin{figure}[htbp]
    \centering
    \includegraphics[width=0.8\textwidth]{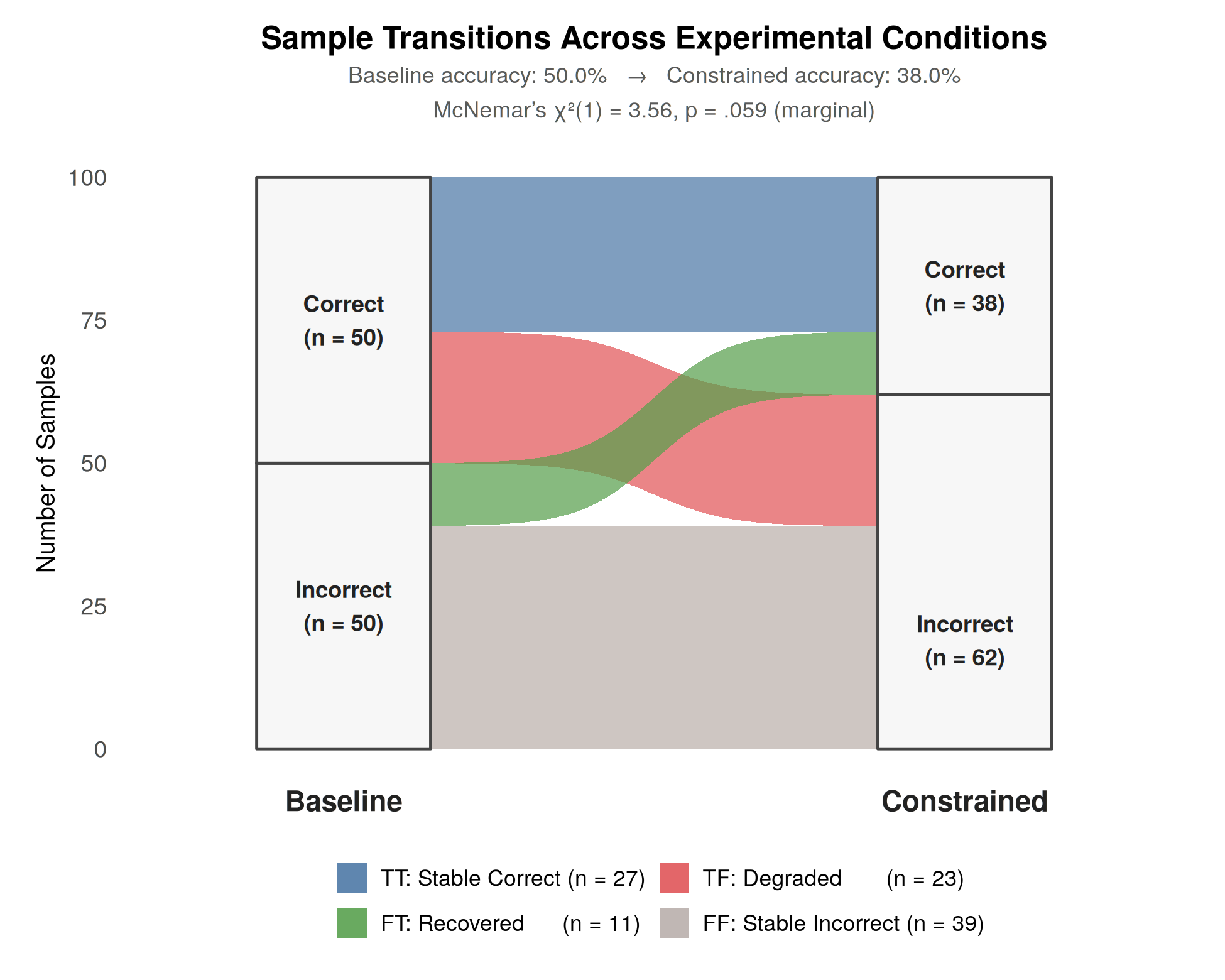}
    \caption{}
    \label{fig:result_alluvial}
    
\end{figure}

The application of logic-guided constrained decoding resulted in a marginal but observable performance penalty. The baseline accuracy of 50.0\% dropped to 38.0\% under constrained conditions. As illustrated by Figure~\ref{fig:result_alluvial}, 23 samples that the model answered correctly in the free-text baseline degraded to incorrect under structural constraints (the TF group). A McNemar’s test ($p \approx 0.059$) confirms this absolute reduction is marginally significant.

This outcome complicates the assumption that ``typed, structured feedback uniformly improves reasoning''. Prior research demonstrates that semi-structured feedback significantly enhances performance when generated by a specially trained external critic \citep{paulREFINERReasoningFeedback2023}. However, when these structural demands are imposed on a single, un-finetuned model via decoding constraints, the intrinsic self-correction process struggles. The model frequently fails to validate its own reasoning without external oracle labels, as observed by \citet{huangLargeLanguageModels2024}. Despite the overall degradation, strict constraints are not entirely detrimental. Eleven samples successfully recovered from an incorrect baseline state to a correct state under constrained decoding (the FT group). Language models tend to commit to early mistakes and generate subsequent justifications, creating a coherence-seeking ``hallucination snowballing'' effect \citep{zhangHowLanguageModel2023}. For the recovered samples, the finite-state machine constraints successfully disrupted this natural left-to-right generation sequence, forcing the model out of incorrect local minima and prompting deliberate belief revision.

\subsection{The Alignment Tax and Cognitive Overload}

\begin{figure}[htbp]
    \centering
    \includegraphics[width=0.8\textwidth]{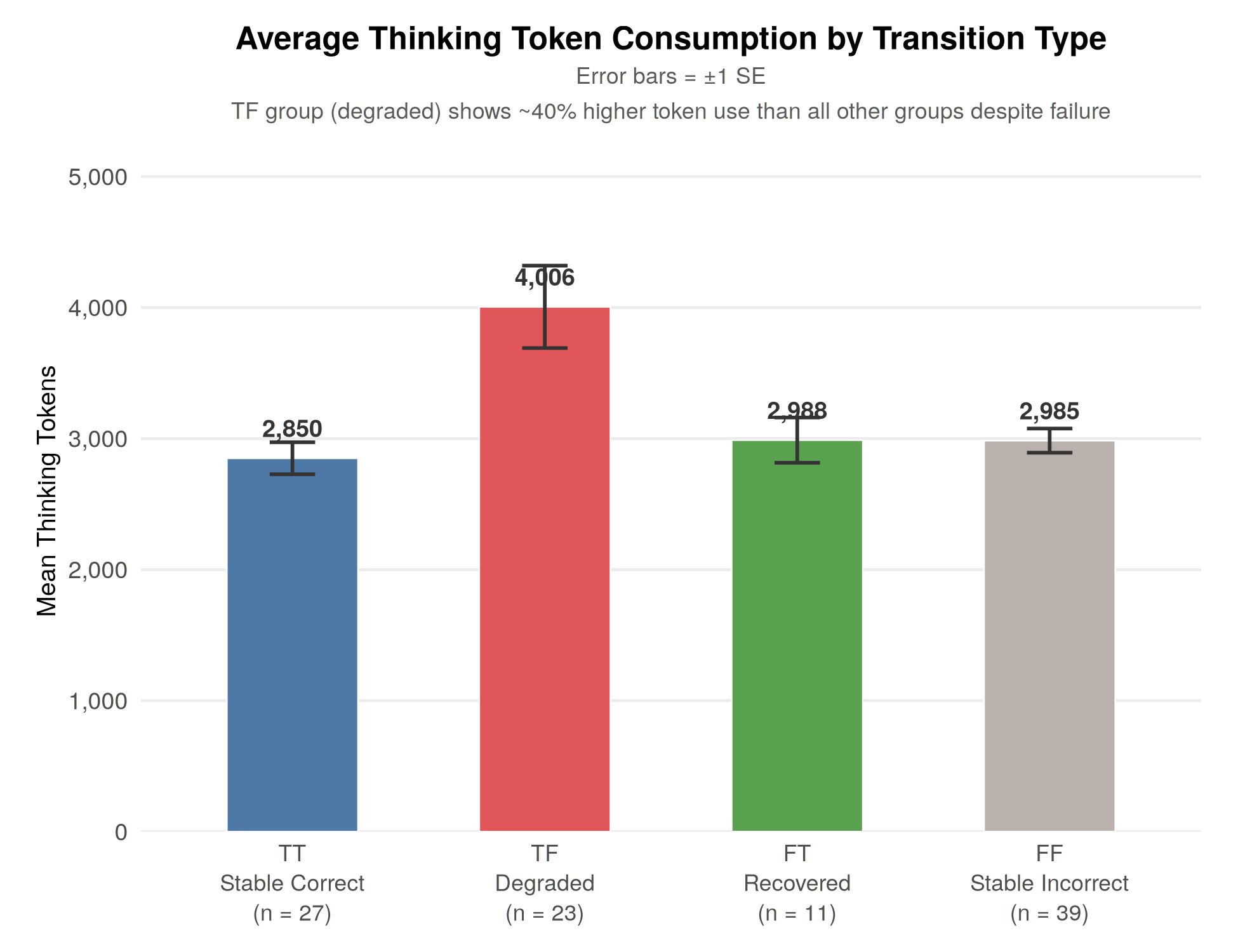}
    \caption{}
    \label{fig:token_count}
    
\end{figure}

The quantitative token data reveals that structural constraints impose a distinctive computational and cognitive overhead. Models that successfully maintained a correct status across both conditions (the TT group) consumed an average of 2,850 thinking tokens. In contrast, the degraded TF group exhibited a massive token consumption surge, averaging 4,005.5 tokens before ultimately failing (Figure~\ref{fig:token_count}). This increasing cost of tokens provides empirical evidence for an ``alignment tax.'' When the decoding library heavily restricts the output vocabulary, the model shifts its attention weights toward syntactic compliance rather than semantic reasoning. Sample 5a8c7595554299585d9e36b6 exemplifies this structural breakdown. The model exhausted its maximum allowance of five trials while outputting consecutive FORMATTING\_MISMATCH errors, with its token consumption climbing linearly from 2,075 to 3,190 tokens.

This behaviour supports the Error Depth Hypothesis, which posits that unguided models default to shallow error correction instead of resolving deep logical flaws \citep{liDecomposingLLMSelfCorrection2025}. The excessive token use indicates cognitive overload; the model thrashes through superficial syntax checks, leaving insufficient computational capacity to execute the necessary multi-hop reasoning.

\subsection{Error Trajectories and the Formatting Trap}

\begin{figure}[htbp]
    \centering
    \includegraphics[width=0.8\textwidth]{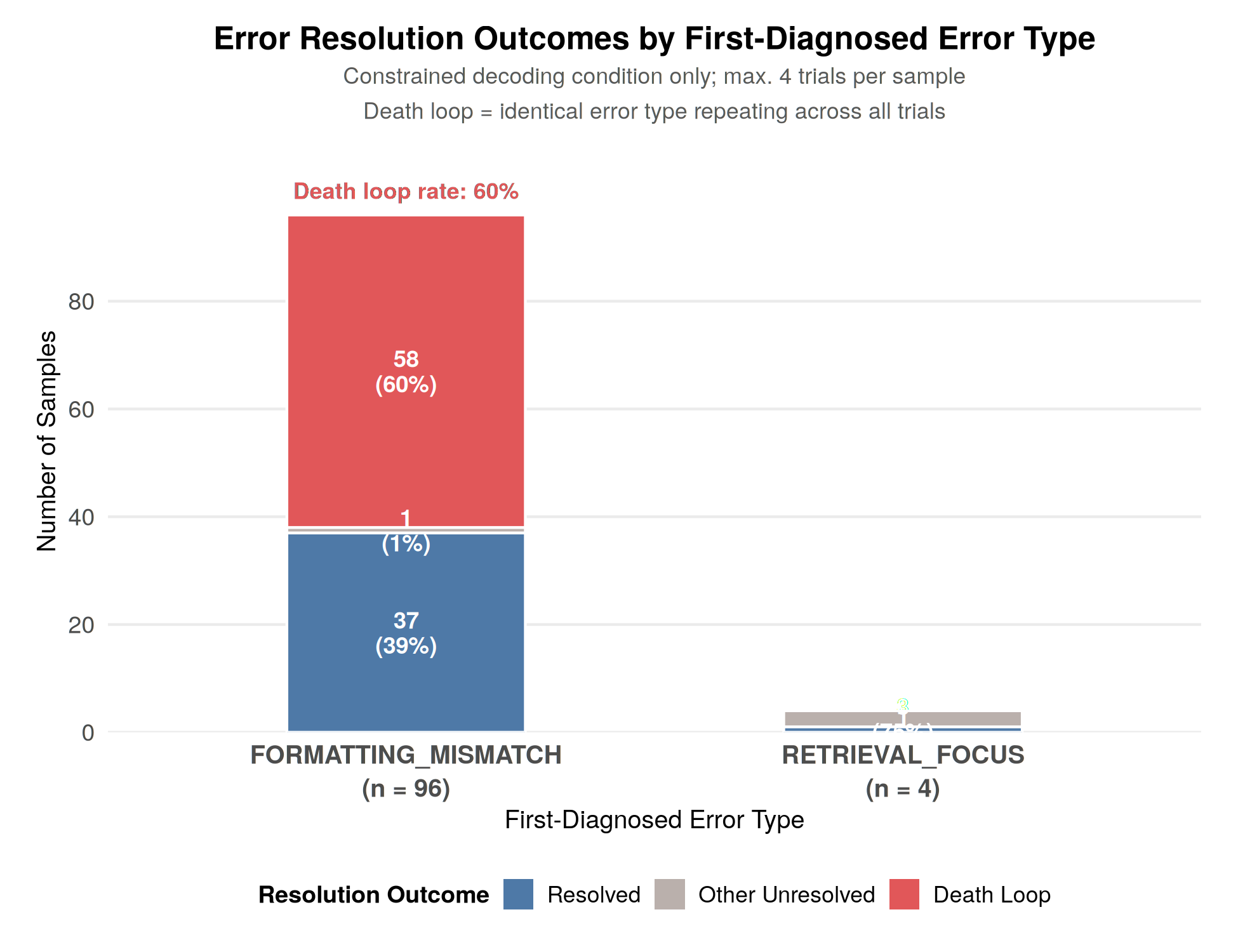}
    \caption{}
    \label{fig:error_reso}
    
\end{figure}


The most significant finding from the reflection trajectories is that structured feedback signals largely failed to guide the model toward identifying retrieval or inference fallacies. Out of 100 first-round diagnoses, 96 categorised the failure as a FORMATTING\_MISMATCH (Figure~\ref{fig:error_reso}). This superficial attribution trapped 58 samples in continuous ``death loops'', where the model repeated the exact same formatting error across all permitted trials.

The qualitative trajectory of sample \texttt{5ac31456554299741d48a1b4} in the Appendix demonstrates how format restrictions mask deeper cognitive errors. The model diagnosed a FORMATTING\_MISMATCH in its first two trials, fully focusing on the format of the response. It only identified the deeper RETRIEVAL\_FOCUS error in the final trial after the syntax perfectly aligned, which then allowed it to extract the correct bridge entity and succeed.

Rather than mitigating hallucination snowballing, the upstream-first attribution strategy inadvertently created ``formatting snowballing.'' Because the evaluation mechanism relies on rigid string matching, surface-level symbol errors trigger immediate failure. The constrained reflector becomes hijacked by these syntax errors and generates repetitive, invalid formatting advice. Without a trained external critic, forcing an 8B-parameter model to adhere to a strict diagnostic schema obstructs deep logical reflection, transforming minor formatting discrepancies into an impassable barrier.

\subsection{Qualitative Analysis: The Syntactic Aligner and the Formatting Twist}
\begin{tcolorbox}[trajectorybox, title=Free Reflection Setting]

\textbf{Question:} Over how many centuries were the "dwelling place of the dead" built?

\textbf{Gold Answer:} three centuries

\medskip
\textbf{Trial 1 Output:}

Thought: The question refers to the Sinagua dwellings at Montezuma Castle. The text states these structures were built over the course of three centuries.

Action: Finish[Three]

\medskip
\textbf{Environment Feedback:} Incorrect

\begin{tcolorbox}[reflectionbox]
\textbf{Reflection:} My answer was correct, but the phrasing discrepancy arose because the context did not explicitly label the Sinagua dwellings as the "dwelling place of the dead." The answer ``Three'' reflects the construction period, but I should better align terminology with the question context.
\end{tcolorbox}

\medskip
\textbf{Trial 2 Output:}

Thought: The dwellings at Montezuma Castle were built between 1100 and 1425 AD, spanning three centuries.

Action: Finish[Three]

\medskip
\textbf{Environment Feedback:} Incorrect

\end{tcolorbox}

\begin{tcolorbox}[trajectorybox, title=Structured Reflection Setting]

\textbf{Question:} Over how many centuries were the "dwelling place of the dead" built?

\textbf{Gold Answer:} three centuries

\medskip
\textbf{Trial 1 Output:}

Thought: The Montezuma Castle text states the dwellings were built over the course of three centuries.

Action: Finish[Three]

\medskip
\textbf{Environment Feedback:} Incorrect

\begin{tcolorbox}[reflectionbox]
\textbf{Error Type:} FORMATTING\_MISMATCH

\textbf{Correction Rule:} Use the exact format: Action: Finish[Three centuries].
\end{tcolorbox}

\medskip
\textbf{Trial 2 Output:}

Thought: The dwellings were built over the course of three centuries.

Action: Finish[Three centuries]

\medskip
\textbf{Environment Feedback:} Correct

\end{tcolorbox}

While the earlier analysis highlighted the pitfalls of structural constraints and the prevalence of the formatting trap, this specific example above illustrates a dialectical twist. It demonstrates the exact scenario where constrained decoding \textbf{succeeds} and resolves shallow alignment failures. 

The example asks the model: ``Over how many centuries were the `dwelling place of the dead' built?''. In the unconstrained baseline setting, the model retrieves the correct context and successfully deduces the numerical answer as ``Three''. The environment rejects this output due to an exact-match evaluation failure. The free-text Reflector then provides a highly accurate semantic diagnosis. It notes that the answer reflects the correct construction period but requires better terminology alignment. Despite this accurate self-diagnosis, the Actor in the subsequent trial fails to execute the correction. It stubbornly repeats the exact same rejected sequence: `Action: Finish[Three]'.

This failure demonstrates a core limitation of unconstrained verbal reflection noted in \citet{shinnReflexionLanguageAgents2023a}. Semantic awareness does not guarantee behavioural execution. The model exhibits a strong coherence-seeking inertia. Without a hard intervention at the decoding layer, the Actor simply falls back into the local minimum of its initial token distribution, which renders the natural language reflection practically useless.

The logic-guided setting presents a noticeable contrast. The Reflector is stripped of its ability to write a nuanced paragraph and is forced into the predefined schema. It categorises the failure as a `FORMATTING\_MISMATCH` and generates a strict, executable correction rule: `Use the exact format: Action: Finish[Three centuries]`. Because the experimental design strictly pairs the error type with a specific correction format, the constraint forces the model to synthesise a deterministic, copy-paste solution rather than a vague semantic observation. When passed to the second trial, this rigid directive successfully overwrites the generation inertia of the model. This intervention results in the correct output and a successful evaluation.
 
The combination of the qualitative analysis and the Error Depth Hypothesis discussed in Section 5.3 provides a more comprehensive image to understand the model's behaviour under the taxonomy's effect. The quantitative data previously revealed that the `FORMATTING\_MISMATCH` category often hijacked the reflector and trapped samples in a loop because it masked deeper cognitive errors. However, this single example proves that when the root cause of the failure genuinely is a superficial string-matching issue, the taxonomy functions as expected. 

Unguided models default to shallow error correction. When forced into a structured taxonomy, the 8B parameter model lacks the cognitive capacity to dig for deep logical flaws, but it becomes exceptionally effective at identifying and patching surface-level syntax discrepancies. This dynamic ultimately exposes a known limitation regarding the strict string-matching requirements of datasets like HotpotQA \cite{yangHotpotQADatasetDiverse2018}. The qualitative evidence suggests that constrained decoding acts as a highly effective syntactic aligner for rigid evaluation metrics, even if it remains a poor semantic depth-probe.
\section{Limitation and Future Work}

A primary limitation of this study lies in the scale of the evaluated model. The experiments rely exclusively on an 8B-parameter architecture. As observed in the qualitative analysis, the restricted intrinsic cognitive capacity of a model at this scale means it often defaults to shallow, syntactic corrections when placed under strict structural constraints. Future work should investigate the scaling laws of constrained self-correction to determine whether larger models, such as those in the 70B parameter class, can use rigid taxonomies for deep logical debugging rather than mere formatting alignment.

Furthermore, the reliance on static exact-match evaluation in established datasets like HotpotQA inherently skews the distribution of failed trials. As baseline models improve, the residual errors in the first trial are increasingly dominated by metric brittleness rather than severe reasoning failures. This dynamic creates a survivor bias in the error pool. We consider this is the main factor heavily favouring superficial string-matching issues. Future evaluations of self-correction mechanisms would benefit from more dynamic, execution-based benchmarks where failures represent definitive logical breakdowns.

While structural constraints successfully mitigate the hallucination snowballing observed in free-text reflection \cite{zhangHowLanguageModel2023}, this study identifies a new vulnerability: structure snowballing. As demonstrated by the ``death loops'' in Section 5.3, strict adherence to a predefined error taxonomy can trap the agent in an unproductive cycle of identical formatting patches. Future research must address this limitation by integrating dynamic fallback mechanisms. Drawing inspiration from search-based reasoning frameworks that allow for state evaluation and backtracking \citep{yaoTreeThoughtsDeliberate2023}, subsequent self-correction architectures should adapt a heuristic trigger. If an agent repeats a constrained correction without achieving environmental success, the system should temporarily lift the decoding constraints. This would allow a brief return to free-text reflection or force a complete rollback to a previous reasoning node, preventing the model from becoming permanently locked in a rigid taxonomy.
\section{Appendix}
\subsection{Result Summary}

\subsubsection{Performance Comparison}

\begin{table}[htbp]
\centering
\small
\begin{tabular}{lcccc}
\hline
Method & Accuracy & Avg Trials & Success@1 & Success@$\leq$2 \\
\hline
Baseline & 0.50 & 0.63 & -- & -- \\
Constrained & \textbf{0.38} & 0.41 & 0.00 & 0.80 \\
\hline
\end{tabular}
\caption{Performance comparison between baseline and constrained reflection.}
\end{table}

\subsubsection{Status Flip Analysis}

\begin{table}[htbp]
\centering
\small
\begin{tabular}{lcc}
\hline
Transition Type & Count & Percentage \\
\hline
Correct $\rightarrow$ Correct & 27 & 27\% \\
Correct $\rightarrow$ Wrong & 23 & 23\% \\
Wrong $\rightarrow$ Correct & 11 & 11\% \\
Wrong $\rightarrow$ Wrong & 39 & 39\% \\
\hline
\end{tabular}
\caption{Status transitions from baseline to constrained setting.}
\end{table}

\subsubsection{Error Type Distribution}

\begin{table}[htbp]
\centering
\small
\begin{tabular}{lcc}
\hline
Error Type & Count & Percentage \\
\hline
Formatting Mismatch & 96 & 96\% \\
Retrieval Focus & 4 & 4\% \\
\hline
\end{tabular}
\caption{Distribution of error types in constrained reflection.}
\end{table}

\subsubsection{Token Usage}

\begin{table}[htbp]
\centering
\small
\begin{tabular}{lccc}
\hline
Method & Avg Tokens & Max Tokens & Std \\
\hline
Constrained & 3183.48 & 8976.25 & 989.11 \\
\hline
\end{tabular}
\caption{Token consumption for constrained reflection.}
\end{table}

\subsection{Case Study}
\label{app:case_study}

To illustrate the qualitative differences between the Free-text Baseline and the Logic-Guided Reflexion (with constrained decoding), we present a representative multi-hop reasoning sample from the HotpotQA dataset. The extremely lenghty action-reflection loops show the ``cogtinive overload'' we aim to discuss in this paper. 

\begin{tcolorbox}[trajectorybox, title={Sample Information}]
\textbf{Question:} Which band has more members, Lawson or American Music Club? \\
\textbf{Sample ID:} \texttt{5ac31456554299741d48a1b4} \\
\textbf{Gold Answer:} Lawson
\end{tcolorbox}

\subsection*{X.1 Free-Text Action Trajectory (Baseline)}
In the standard Reflexion baseline, the actor produces free-text thoughts and actions, and the reflector provides unstructured feedback. The agent successfully answered this question in 2 trials.

\begin{tcolorbox}[trajectorybox, title={Baseline Trajectory}]
\textbf{Trial 1 Actor Output:}

{\itshape
Okay, let's tackle this question. The user is asking which band has more members: Lawson or American Music Club. First, I need to look at the relevant context provided. Let me scan through the texts again.Starting with Lawson: The entry for Lawson mentions they are an English pop rock band consisting of four members—Andy Brown, Ryan Fletcher, Joel Peat, and Adam Pitts. So that's four members total.Next, American Music Club: The text says it's an American, San Francisco-based indie rock band led by Mark Eitzel. But it doesn't explicitly state the number of members. Wait, there's another mention of American Music Club in the Atwater Afternoon entry, which says the band members other than Mark Eitzel wrote some songs. But that doesn't specify the total number. However, in the Mean Mark Eitzel Gets Fat entry, it's mentioned that the album includes songs by members of the band other than Mark Eitzel. But again, no numbers. Wait, maybe I missed something. Let me check again. The American Music Club's entry just says it's led by Mark Eitzel but doesn't list other members. However, in the context about the Atwater Afternoon, it mentions that two of the original songs were written by members other than Mark Eitzel. That implies there are other members, but how many? The original question is about the number of members. Wait, maybe there's another part. The One Star Hotel entry mentions Steve Yutzy-Burkey as the frontman, but that's a different band. The Idaho band has Jeff Martin and John K. Berry as founders, but again, different band. Wait, maybe the answer is that American Music Club has more members because Lawson has four, and American Music Club is mentioned as having more than just Mark Eitzel. But how many exactly? The context doesn't specify. Wait, maybe I need to infer. For example, in the Atwater Afternoon entry, it says \"the band\" which includes Mark Eitzel and others. But without explicit numbers, maybe the answer is that American Music Club has more members. But how can I be sure? Alternatively, maybe the answer is that Lawson has four members, and American Music Club has more. But since the context doesn't state the exact number for American Music Club, perhaps the answer is that American Music Club has more members. Wait, but the user might be expecting that American Music Club has more. Wait, but the context for American Music Club only mentions Mark Eitzel as the leader. Maybe the answer is that Lawson has four members, and American Music Club has more. But without explicit numbers, maybe the answer is that American Music Club has more. Alternatively, maybe the answer is that American Music Club has more members. Wait, perhaps the answer is that American Music Club has more members. Because the context for American Music Club mentions that other members wrote songs, implying there are other members. But how many? The question is about which has more members. If Lawson has four, and American Music Club has at least two (Mark Eitzel plus others), but maybe more. However, without exact numbers, maybe the answer is that American Music Club has more. But I'm not sure. Wait, maybe the answer is that American Music Club has more members. But I need to check again. Wait, the context for American Music Club says it's led by Mark Eitzel. The Atwater Afternoon entry mentions that two of the songs were written by members other than Mark Eitzel. So that implies there are at least two other members. So total members would be Mark Eitzel plus at least two others, making three. But Lawson has four. So then Lawson would have more members. Wait, but maybe there are more members. Alternatively, maybe the answer is that American Music Club has more members. But I'm not sure. Wait, the context for American Music Club doesn't specify the number of members. The Lawson entry clearly states four members. So if American Music Club has more than four, then it's the answer. But if it's only three, then Lawson has more. But the context doesn't say. Wait, maybe the answer is that American Music Club has more members. Because the question is about which band has more members. Since the context for American Music Club mentions that other members wrote songs, implying there are other members besides Mark Eitzel. But without exact numbers, perhaps the answer is that American Music Club has more members. But I'm not certain. Alternatively, maybe the answer is that Lawson has four members, and American Music Club has more. But without explicit numbers, maybe the answer is that American Music Club has more. But I'm not sure. Wait, maybe the answer is that American Music Club has more members. But I need to make a decision based on the given context. Wait, the context for American Music}

\medskip
\textbf{Trial 1 Environment Feedback:} \\
\texttt{System Log: INCORRECT.} \\

\textbf{Trial 1 Reflector Output (Free-text):}

{\itshape
Okay, let's see. The user asked which band has more members: Lawson or American Music Club. First, I need to check the context provided. For Lawson, the text clearly states they have four members: Andy Brown, Ryan Fletcher, Joel Peat, and Adam Pitts. That's straightforward.Now, for American Music Club, the context mentions it's led by Mark Eitzel but doesn't list the number of members. However, there's a mention in the Atwater Afternoon entry that two songs were written by members other than Mark Eitzel. This implies there are at least two other members, making the total at least three. But the context doesn't specify if there are more. Wait, but the question is about which band has more members. Since Lawson has four, and American Music Club is at least three, but maybe more. However, the context doesn't give exact numbers for American Music Club. So, based on the given information, the answer should be Lawson has more members. But the user might have expected American Music Club to have more. Wait, maybe I missed something. Let me check again. The American Music Club's entry says it's a band led by Mark Eitzel. The Atwater Afternoon entry mentions that two songs were written by other members, so there are at least two others. So total members would be Mark Eitzel plus at least two, making three. Therefore, Lawson has four, which is more. But the user's answer might have been different. Wait, the user's previous trial might have had an error. Wait, the user's reflection here is about their own previous trial. Wait, the user provided a previous trial where they might have answered incorrectly. In the previous trial, the user's answer was incorrect. The correct answer is Lawson has four members, and American Music Club has three (Mark Eitzel plus two others). Therefore, Lawson has more. But if the user answered American Music Club, that would be wrong. So the reflection should point out that the user might have confused the number of members, assuming American Music Club has more without explicit information. The correct answer is Lawson. The user's mistake was not checking the exact numbers and assuming based on the mention of other members. So the diagnosis is that the user might have assumed American Music Club has more members without explicit data, leading to an incorrect answer. The plan is to carefully count the members mentioned in the context and avoid assumptions.</think>Diagnosis: The error likely stemmed from assuming American Music Club has more members based on the mention of other contributors, without explicitly counting them. The context only states Lawson has four members, while American Music Club’s member count is implied (at least Mark Eitzel plus others) but not specified. Assuming a higher number for American Music Club without explicit data led to an incorrect answer.  New Plan: Carefully extract and count all explicitly mentioned members for each band. For American Music Club, note that the context only references Mark Eitzel as the leader and others contributing songs, but does not list total members. Thus, Lawson (4 members) has more than American Music Club (at least 3, but unspecified). Avoid assumptions; rely strictly on explicit numerical data.}

\medskip
\textbf{Diagnosis:} The error stemmed from inferring that American Music Club had more members without sufficient textual support. \\
\textbf{New Plan:} Count only the explicitly supported members for each band and avoid unsupported extrapolation. \\

\textbf{Trial 2 Actor Output:} \\

{\itshape
Okay, let's tackle this question. The user is asking which band has more members: Lawson or American Music Club. First, I need to check the context provided for each band. Starting with Lawson: The context clearly states that Lawson is an English pop rock band consisting of four members—Andy Brown, Ryan Fletcher, Joel Peat, and Adam Pitts. That's straightforward. So Lawson has exactly four members.Now, looking at American Music Club. The context mentions that it's an American, San Francisco-based indie rock band led by Mark Eitzel. However, there's no explicit mention of the total number of members. But there's a mention in the Atwater Afternoon entry that two of the original songs on the album were written by members other than Mark Eitzel. This implies there are at least two other members besides Mark Eitzel. However, the context doesn't specify if there are more members beyond those two. Wait, but the Undertow Orchestra entry mentions Mark Eitzel as part of that group, but that's a different band. The key point here is that American Music Club's member count isn't directly stated. The only information is that Mark Eitzel is the leader, and there are at least two other members. So the minimum number of members for American Music Club is three (Mark Eitzel plus two others). However, the context doesn't confirm if there are more members. Comparing the two, Lawson has four members explicitly listed. American Music Club's member count is at least three but not confirmed to be more. Therefore, based on the given information, Lawson has more members than American Music Club. The user might have thought American Music Club has more because of the mention of other contributors, but without explicit numbers, we can't assume. The correct answer is Lawson.</think>Thought: The question asks which band has more members. The context explicitly states Lawson has four members. For American Music Club, while Mark Eitzel is the leader and two other members contributed songs, the total member count is not specified. Since Lawson's number is clearly stated (4) and American Music Club's is implied to be at least 3 but unspecified, Lawson has more members. 
}

\textit{[...]} \texttt{Action: Finish[Lawson]} \\
\textbf{Environment Feedback:} \texttt{System Log: Success.}
\end{tcolorbox}

\subsection*{X.2 Constrained-Text Action Trajectory (Logic-Guided Reflexion)}
Under the treatment condition, the Outlines constrained decoding forces the reflector to strictly output a predefined schema: \texttt{Error(Type, Correction\_Rule)}. This sample required 3 trials to succeed, demonstrating how structural constraints handle intermediate failures like \texttt{FORMATTING\_MISMATCH} and \texttt{RETRIEVAL\_FOCUS}.

\begin{tcolorbox}[trajectorybox, title={Logic-Guided Reflexion Trajectory (Treatment)}]



\begin{tcolorbox}[reflectionbox]
\textbf{Trial 1 Structured Reflection:}

\textbf{Error Type:} \texttt{FORMATTING\_MISMATCH}

\textbf{Correction Rule:} You did not provide a valid \texttt{Action: Finish[...]} format in your output. Next time, use either \texttt{Action: Finish[more members]} or \texttt{Action: Finish[Lawson]}, depending on the correct answer.

Next time, ensure you strictly follow the required format by enclosing your final answer exactly as \texttt{Action: Finish[more members]} or Action: \texttt{Finish[Lawson]} on the correct answer.
\end{tcolorbox}

\vspace{0.5em}
\textbf{Trial 2 Actor Output:}

I must ensure I follow the correct format by providing a valid \texttt{Action: Finish[...] } the answer. Based on the context, Lawson has four members, while American Music Club's exact number isn't specified but implies more than one. However, since the question asks which has more members, and Lawson is explicitly stated to have four, while American Music Club's member count isn't clearly stated, the answer is Lawson

\texttt{Action: Finish[Lawson]}
\texttt{Extracted Answer: ...}

\textbf{Trial 2 Environment Feedback:} \texttt{System Log: Fail.}

\textbf{Trial 3 Structured Reflection:}

\textbf{Error Type:} \texttt{RETRIEVAL\_FOCUS}

\textbf{Correction Rule:} You focused on the explicit member count of Lawson (4 members) but failed to properly infer the member count of American Music Club. The context states that American Music Club is led by Mark Eitzel and includes other members who wrote songs, implying a minimum of 2-3 members. Therefore, American Music Club likely has more members than Lawson. Next time, carefully infer the member count of bands based on contextual clues, not just explicit numbers.





\vspace{0.5em}
\textbf{Trial 3 Actor Output:}

I must carefully infer the member count of American Music Club based on contextual clues. The context states that American Music Club is led by Mark Eitzel and that other members wrote songs, implying at least two other members. This suggests a minimum of three members. Since Lawson has four members, the correct answer is Lawson. However, the previous reflection indicated that American Music Club likely has more members. I need to re-examine the context for any explicit mentions of member counts. The American Music Club entry only mentions Mark Eitzel as the leader, but the Atwater Afternoon entry mentions that two songs were written by members other than Mark Eitzel, implying at least two other members. Thus, American Music Club has at least three members, while Lawson has four. Therefore, Lawson has more members.

\texttt{Action: Finish[Lawson]}

\textbf{Trial 3 Environment Feedback:} \texttt{System Log: Success.}
\end{tcolorbox}

\bibliographystyle{plainnat}
\bibliography{agentic_llm}

\end{document}